\documentclass[conference]{IEEEtran}
\IEEEoverridecommandlockouts
\usepackage{booktabs}
\usepackage{multirow}
\usepackage{tabularx}
\usepackage{graphicx}
\usepackage{arydshln} 
\usepackage{amsmath}
\usepackage{amsfonts}
\usepackage{bm}%加粗
\usepackage{mathrsfs}
\usepackage[caption=false,font=scriptsize,labelfont=rm,textfont=rm]{subfig}
\usepackage{makecell}

\usepackage{cite}
\usepackage{amsmath,amssymb,amsfonts}
\usepackage{algorithmic}
\usepackage{graphicx}
\usepackage{textcomp}
\usepackage{xcolor}
\def\BibTeX{{\rm B\kern-.05em{\sc i\kern-.025em b}\kern-.08em
    T\kern-.1667em\lower.7ex\hbox{E}\kern-.125emX}}
\begin{document}

\title{ITERTL: An Iterative Framework for Fine-tuning LLMs for RTL Code Generation\\
}

\author{\IEEEauthorblockN{
    Peiyang Wu,
    Nan Guo,
    Xiao Xiao,
    Wenming Li,
    Xiaochun Ye,
    Dongrui Fan}
\IEEEauthorblockA{\textit{Institute of Computing Technology} \\
\textit{Chinese Academy of Sciences}\\
Beijing, China \\
\{wupeiyang,guonan\}@ict.ac.cn}
}
% \and
% \and
% \and
% \and
% \and

\maketitle

\begin{abstract}
Recently, large language models (LLMs) have demonstrated excellent performance, inspiring researchers to explore their use in automating register transfer level (RTL) code generation and improving hardware design efficiency. However, the existing approaches to fine-tune LLMs for RTL generation typically are conducted on fixed datasets, which do not fully stimulate the capability of LLMs and require large amounts of reference data, which are costly to acquire. To mitigate these issues, we innovatively introduce an iterative training paradigm named ITERTL. During each iteration, samples are drawn from the model trained in the previous cycle. Then these new samples are employed for training in current loop.  Furthermore, we introduce a plug-and-play data filtering strategy, thereby encouraging the model to generate high-quality, self-contained code.  Our model outperforms GPT4 and state-of-the-art (SOTA) open-source models, achieving remarkable 53.8\% pass@1 rate on VerilogEval-human benchmark. Under similar conditions of data quantity and quality, our approach significantly outperforms the baseline. Extensive experiments validate the effectiveness of the proposed method.
\end{abstract}


\section{Introduction}

Manually writing hardware description language (HDL) code (e.g., Verilog) is an essential part of the current hardware design process. This step is labor-intensive and prone to errors, consuming a significant amount of developers' time. As LLMs have demonstrated excellent performance in natural language processing and code generation, researchers are exploring their potential as an aid in hardware design \cite{chang2023chipgpt,blocklove2023chip,thakur2023benchmarking,liu2023verilogeval,liu2023chipnemo}.

\begin{figure}[t]
  \centering
  \includegraphics[width=.7\linewidth]{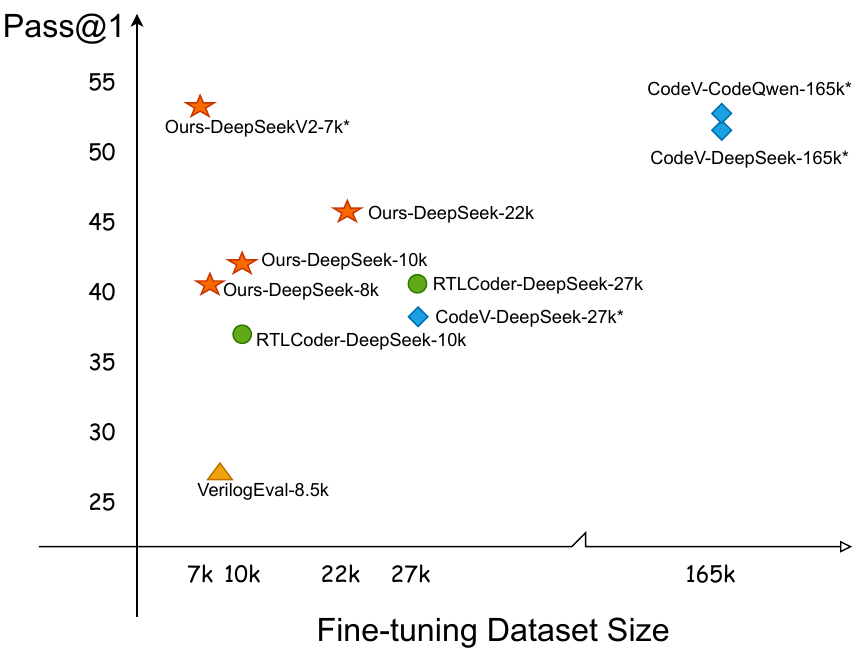}
  \caption{Fine-tuning dataset size and pass@1 on VerilogEval-human. Our method achieves excellent results relying on limited data. We use * to denote models trained using higher quality data compared with RTLCoder-27k\cite{liu2023rtlcoder}.
  \iffalse Our datasets are derived from subsets of RTLCoder-27k\cite{liu2023rtlcoder}.\fi}
  \label{fig:datasize}
%\vspace{-10pt}
  
\end{figure}

% \begin{figure}[t]
%   \centering
%   \includesvg[width=\linewidth]{formal-datasize.drawio.svg}
%   \caption{Fine-tuning dataset size and pass@1 on VerilogEval-human. Our method achieves excellent results relying on limited data. Our datasets are derived from subsets of RTLCoder-27k\cite{liu2023rtlcoder}.}
%   \label{fig:datasize}
  
% \end{figure}

In order to get more professional LLMs in RTL code generation, a common method is to use domain-specific datasets to fine-tune LLMs. VeriGen \cite{thakur2023benchmarking} leverages corpora from GitHub and Verilog textbooks to fine-tune open source LLMs. ChipNeMo \cite{liu2023chipnemo} customizes LLaMa2 \cite{touvron2023llama} for functionalities such as chatbot, generating EDA tool script, and bug summarization. RTLCoder \cite{liu2023rtlcoder} develops an automated flow to generate instruction-code pairs  and uses them to fine-tune LLMs leveraging code quality assessment. CodeV\cite{zhao2024codev} constructs a fine-tuning dataset by generating multi-level summaries of Verilog code utilizing GPT-3.5. \iffalse MEV-LLM \cite{nadimi2024multi} proposes to fine-tune corresponding LLMs for hardware designs of varying complexity and integrate them together to boost performance.\fi BetterV\cite{pei2024betterv} introduces generative discriminators to guide the optimization of Verilog code. However, most existing methods follow the conventional paradigm of deep learning,  which involves initially gathering data and then training the model. This approach may lead to two negative effects:\textbf{(1)}Since the model is trained on a limited amount of collected data, the space available for the model to explore is thus constrained, resulting in a narrow coverage of feedback signals, which can be automatically obtained through RTL tools.\textbf{(2)}There is a mismatch between the distributions of training samples and the LLM under training, which can lead to estimation errors \cite{li2023policy,liu2023statistical} in optimization process. \iffalse Specifically, if the training samples are not directly generated by the LLM intended for training, their distributions are obviously misaligned. However, even if the training samples are generated by the LLM at the beginning (e.g. the pretrained model), their distributions will still mismatch because the distribution of the LLM is shifting during the training process.\fi Due to the aforementioned  negative effects, the generation capability of existing fine-tuned models are limited. Additionally, a large number of reference samples are required for fine-tuning. However, in the field of hardware design, particularly for tasks involving Verilog code generation, the acquirance of training data is very expensive for reasons such as intellectual property protection. 

% \iffalse Therefore, it is crucial to develop training methods that can perform well with small datasets to effectively enhance the fine-tuning of LLMs.\fi In order to foster exploration and mitigate the distribution deviation, an intuitive idea is to introduce reinforcement learning (RL) methods to allow LLMs to refine their policies through the interaction with the environment \cite{ouyang2022training,stiennon2020learning,le2022coderl,liu2023rltf}. However, it is well-known that the process of RL is often complex, resource-intensive, and unstable. Typically, these RL algorithms need to maintain several models at the same time (such as policy model, value model and reference model in PPO algorithm \cite{schulman2017proximal}) and conduct extensive hyper-parameters tuning, which will incur significant computational overhead and pose substantial obstacles for users.

To tackle the issues mentioned, we introduce an iterative training scheme to fine-tune LLMs specifically for generating Verilog code. Unlike previous methods learning from one fixed training set, our approach iteratively updates the training samples and automatically assesses them using RTL tools during the training process. Besides, to improve the quality of training samples, we develop a data filtering strategy based on code structure to effectively encourage LLMs to generate high-quality, self-contained RTL code, which means it doesn't need to instantiate other submodules. This strategy is conducted by removing samples whose code contains multiple modules or assigning them lower scores. By integrating this data filtering strategy with the iterative training paradigm, the capability of LLMs is substantially improved. \iffalse Since the training samples are iteratively sampled from the updated model, the mismatch between distributions of the training samples and LLMs, as well as the resulting estimation error, are mitigated accordingly. As the first method which iteratively updates training samples in the RTL code generation field, our method enables the model to achieve SOTA performance. Moreover, our approach, which requires minimal hyper-parameter tuning without handling multiple models during training, is quite easy to implement.\fi

Our contributions can be outlined as follows:

\begin{itemize}
% \item We develop an iterative supervised fine-tuning scheme for training LLMs to generate Verilog code. By expanding the exploration scope of models and reduces estimation errors caused by distribution mismatches, this solution can effectively boost the Verilog code generation capability. To the best of our knowledge, this is the first sampling-training iterative fine-tuning scheme proposed in this field. \iffalse and substantially decreases the number of externally-sourced reference samples needed for training. \fi
% %\vspace{-8pt}
\item To the best of our knowledge, we are the first to propose a sampling-training iterative fine-tuning scheme utilizing the feedback from the RTL tool to tailor LLMs for the Electronic Design Automation (EDA) field. This approach not only effectively enhances the functional correctness of RTL code but also holds broad potential for applications in other areas where automatic feedback is available. \iffalse By expanding the exploration scope of models and reduces estimation errors caused by distribution mismatches, this solution can effectively boost the Verilog code generation capability. To the best of our knowledge, this is the first sampling-training iterative fine-tuning scheme proposed in this field.  and substantially decreases the number of externally-sourced reference samples needed for training. \fi

\item We introduce a plug-and-play data filtering strategy, which effectively steer LLMs towards generating high-quality, self-contained module and can be either used independently or combined with the iterative training paradigm to further boost performance.

\item Our models achieve excellent results relying on limited data. As shown in Figure~\ref{fig:datasize}, when controlling for comparable data quality and quantity, our models significantly outperform the baseline models, achieving state-of-the-art (SOTA) performance. 

\end{itemize}

\section{Approach}
In this section, we firstly provide an overview of the proposed framework, followed by detailed explanations of the implementation of the iterative fine-tuning and the data filter, which are our key innovative designs.
% In this section, we firstly detail the workflow of our proposed approach. Then analyze the deficiencies of related methods and reveal the improvements of our approach from a theoretical viewpoint.

%\vspace{-5pt}
\subsection{Framework Overview}\label{sec:frame}

\begin{figure}[!h]
  \centering
  \includegraphics[width=.9\linewidth]{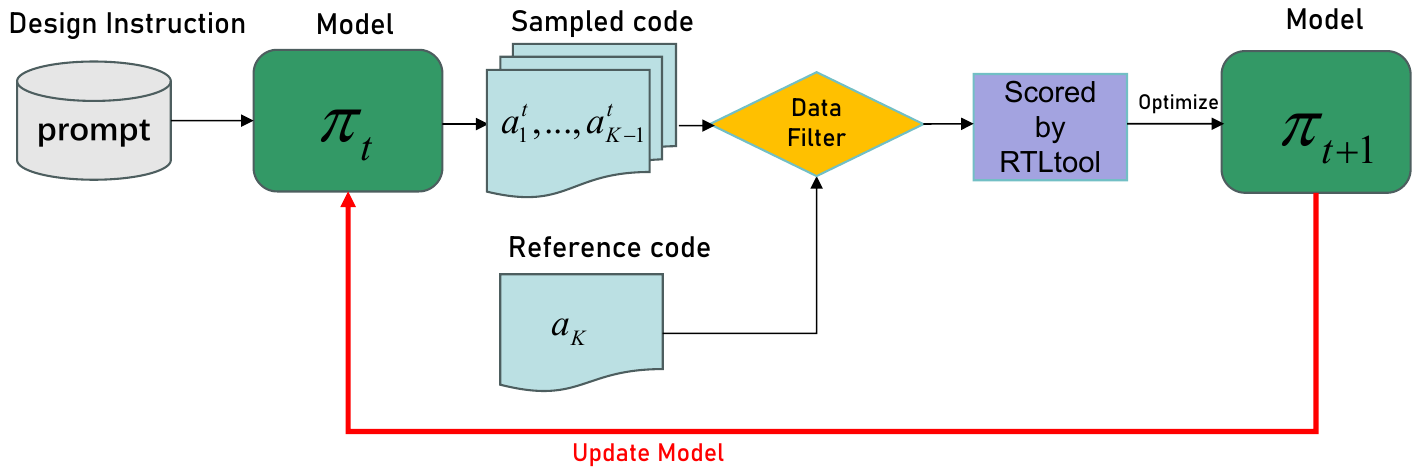}
  \caption{Our proposed ITERTL, which further introduces the iterative training paradigm and the data filter to significantly enhance the model's capability.}
\label{fig:Comparisons}
\end{figure}

Figure~\ref{fig:Comparisons} illustrates the workflow of our approach. The model takes design instructions as input prompts and generates sample code, which is fed into the data filter for preprocessing, along with reference code produced by a teacher model (e.g. GPT-3.5).  The filtered data is then scored by an RTL tool and utilized to optimize the model. After converging during this training cycle, the model will be subsequently employed in the next cycle  to update the sample code.

% \begin{figure}[!htb]
% \centering

% % \hfill
% \begin{minipage}[b]{.9\linewidth}
%     \centering
%     \subfloat[][ITERTL: Iterative finetuning with the help of the data filter (ours)]{\label{ITERTL: Iterative finetuning with the help of the data filter}\includegraphics[width=\textwidth]{3.pdf}}

% \caption{\textbf{Comparisons of different finetuning methods.} (a) represents the vanilla finetuning approach which employs reference instruction-code pairs, e.g. VerilogEval \cite{liu2023verilogeval}. (b) indicates the method of introducing sampled data, e.g. RTL-Coder \cite{liu2023rtlcoder}. (c) is our proposed ITERTL, which further introduces the iterative training paradigm and the data filter to significantly enhance the model's capability. }
% \label{fig:Comparisons}
% %\vspace{-12pt}
% \end{figure}

%\vspace{-5pt}

\subsection{Iterative Fine-tuning}\label{sec:Iterative}

\iffalse We briefly introduce the procedure in a single loop at first, and then explain the iteration process.  during a single loop, our method is similar to  RTLCoder \cite{liu2023rtlcoder}, which has been proven to effectively enhance the model's performance.\fi Building upon the non-iterative approach RTLCoder \cite{liu2023rtlcoder}, we further extend it into a sampling-training iterative framework. As shown in Figure~\ref{fig:Comparisons}, in the $t$-th round of training, for each input instruction, denoted as $s$, there are $K$ corresponding output responses $a_{k}^{t}$, $1\le k\le K$. Among these, the first $K-1$ responses are sampled from the model $\pi _{t} $ acquired from previous training iteration. While the last response $a_{K}^{t}$, which serves as the reference data, can be obtained from another teacher LLM or human annotation. For simplicity, we generate $a_{K}^{t}$ only once and omit the superscript $t$ from $a_{K}^{t}$. Following RTLCoder \cite{liu2023rtlcoder}, each response $a_{k}^{t}$ is assigned a quality score $z_{k}^{t}$ ranging from 0 to 1, by utilizing the open-source RTL tool Pyverilog \cite{takamaeda2015pyverilog} to verify the syntactic correctness and Rouge-L metric to measure the similarity, as Equation~\ref{eqn:score} shows. \iffalse This process can be formalized as follows: This process can be formalized as Equation~\ref{eqn:score} shows.\fi However, we want to emphasize that our framework can be further extended when the feedback from other RTL tools (like PPA) is available.
% Specifically, Pyverilog is firstly used  to verify the syntactic correctness of RTL code. If the syntax is correct, then the score is set to 1. If incorrect, the score is set to a value between 0 and 1, determined by the Rouge-L metric between sampled response $a_{k}^{t}$ and reference data $a_{K}$.

\begin{equation}\label{eqn:score}
z_{k}^{t} = 
\begin{cases} 
1 & \text{if } a_{k}^{t} \text{ is syntactically correct} \\
\text{Rouge-L}(a_{k}^{t}, a_K) & \text{otherwise}
\end{cases}
\end{equation}

% The loss function comprises two components: the ranking loss \cite{liu2023rtlcoder,yuan2023rrhf,liu2022brio} and the Maximum Likelihood Estimation (MLE) loss. 

Then the conditional log probability (length-normalized) is calculated by leveraging the model $\pi$ that is being trained:

\begin{equation}\label{eqn:log_p}
p_{k}=\frac{\sum_{j} \log P_{\pi}\left(a_{k, j}^{t} \mid s, a_{k,<j}^{t}\right)}{\left\|a_{k}^{t}\right\|}
\end{equation}

Combining the quality score $z$ and log probability $p$, the ranking loss \cite{liu2023rtlcoder,yuan2023rrhf,liu2022brio} can be computed as shown in Equation~\ref{eqn:ranking}. Unlike the previous work \cite{liu2023rtlcoder}, here we discard the use of softmax normalization to the log probability $p$ to avoid possible vanishing gradients brought by softmax saturation.

\begin{equation}\label{eqn:ranking}
L_{ranking}^{t}=\sum_{z_{k}^{t}<z_{\tau}^{t}-\beta} \max \left(p_{ k}-p_{\tau}+\alpha, 0\right)
\end{equation}

MLE loss component is the common cross entropy loss relative to the reference sample $a_{K}$:

\begin{equation}\label{eqn:ce}
L_{ce}=-\sum_{j} \log P_{\pi}\left(a_{K, j} \mid s, a_{K,<j}\right)
\end{equation}

The overall loss function is a weighted sum of both parts:

% %\vspace{-5pt}
\begin{equation}\label{eqn:total_loss}
L^{t}=L_{ce}+\lambda L_{ranking}^{t}
\end{equation}
% %\vspace{-5pt}
% To control the scale relationship between the two loss functions, we set $\lambda$ equal to $sg(L_{ce})$, where $sg(\cdot )$ represents a stop-gradient operation.

After the initial round of training, we obtained the converged model $\pi _{t+1} $. Even though the capability of model $\pi _{t+1} $ has improved comparing to $\pi _{t} $, there remains room for further enhancement. So we propose to resample new responses $a_{k}^{t+1}$ using the new model $\pi _{t+1} $ and assess new quality scores $z_{k}^{t+1}$ automatically by RTL tools. Resampling can bring new feedback as the model's distribution has shifted after previous round of training. We compute the loss function $L^{t+1}$ and optimize the model's parameters with new data. This process is repeated until the training process concludes.

\subsection{Data Filter}\label{sec:datafilter}
\begin{figure}[!htb]
\begin{minipage}[t]{.48\linewidth}
    \centering
    \subfloat[][Incorrect code]{\label{incorrect, non-self-contained code}\includegraphics[width=\textwidth]{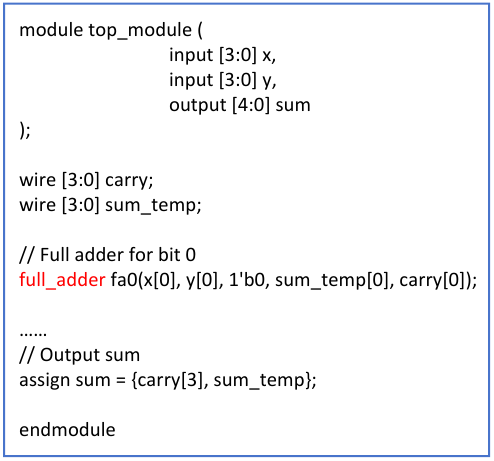}}

\end{minipage} %\par
% \medskip
% \\
\begin{minipage}[t]{.48\linewidth}
    \centering

    \subfloat[][Correct code]{\label{Finetune with reference data and sampled data}\includegraphics[width=\textwidth]{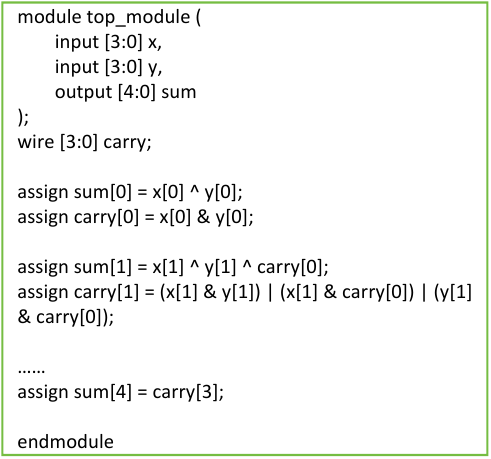}}
\end{minipage} %\par

\caption{Examples of output code for a 4-bit adder. (a) shows the model  trained directly
with unfiltered data generating incomplete code without the implementation of the submodule. (b) shows the model trained with filtered data outputing correct implementation.}
\label{fig:badcode}
\end{figure}

We examine the output code of models trained directly with unfiltered data and find that these models tend to produce incorrect, non-self-contained code. As Figure~\ref{fig:badcode} (a) shows, models trained with unfiltered data output incomplete code for a 4-bit adder without the implementation of the submodule. 

Given the current limiting ability of LLMs to generate RTL code, we believe the primary goal at this stage should first be to improve their ability to generate correct self-contained code, which does not  instantiate other submodules. As for more complex code, it can be composed of many independent modules. To encourage the model to generate high-quality self-contained code, we adopt a two-pronged data filtering strategy:

(1) For reference code, we directly discard samples that contain multiple modules. (2) For sampled code, we fix the score $z_{k}^{t}$ of the sample containing multiple modules at -1, making it lower than the scores of normal samples, which are between 0 and 1. As shown in Figure~\ref{fig:badcode} (b), fine-tuning with the filtered data discourages the model from generating non-self-contained code, resulting in correct outputs.

% As a result, the model will be discouraged from generating non-self-contained code.

% As Figure~\ref{fig:badcode} (b) shows, after fine-tuned with the filtered data, the model can output correct and self-contained code.

% \captionsetup[subfloat]{% figure 也可以
%     font=bf, % 设置标题的字体为粗体
%     labelfont=small, % 设置标签的字体为小字号
%     format=hang, % 设置标题格式为悬挂缩进
%     indention=1cm, % 设置标题缩进为1cm
%     labelsep=period, % 设置标签和标题的分隔符为句号
%     justification=raggedright, % 设置标题的对齐方式为左对齐
%     singlelinecheck=false, % 设置多行标题的对齐方式为左对齐
%     width=0.8\textwidth, % 设置标题的宽度为文本宽度的80%
%     position=bottom, % 设置标题位置在底部
%     skip=10pt, % 设置标题和内容之间的垂直距离为10pt
% }

%\vspace{-5pt}
% \begin{figure}[!h]
%   \centering
%   \includegraphics[width=.6\linewidth]{bad_code_comp2.pdf}
%   \caption{Examples of erroneous output code from the model trained directly with unfiltered data. The model generates incorrect code which lacks the implementation of submodule for an full adder. The submodule is highlighted in red to indicate the non-self-contained characteristics.}
% \label{fig:badcode}
% \end{figure}

%\vspace{-20pt}

\section{Experimental Evaluation}\label{sec:exp}

\subsection{Experimental Setup}
\textbf{Benchmark:} We choose a comprehensive evaluation dataset, VerilogEval \cite{liu2023verilogeval}, as the benchmark to measure the performance. \iffalse This dataset consists of diverse Verilog tasks ranging from simple to complex, such as combinational circuits, finite state machines, code debugging, constructing testbenches, and so on.\fi Two sets of design instructions are provided: the first one is generated by LLM, named \textbf{VerilogEval-machine}; the other one is manually written, named \textbf{VerilogEval-human}. Corresponding golden solutions are also provided to verify functional correctness.

\begin{table*}[t]
\centering
  \caption{Performance on VerilogEval Benchmark\cite{liu2023verilogeval}. \iffalse The results of GPT-3.5, GPT-4, VerilogEval, Codegen and VeriGen come from \cite{liu2023verilogeval}. The results of RTLCoder come from \cite{liu2023rtlcoder}.\fi \textbf{The results are divided into 4 groups based on the size and quality of the fine-tuning dataset. \iffalse The solid line is used to indicate whether models have reported a clear amount of fine-tuning data.\fi} We use * to mark models that utilized higher quality training data compared to RTLCoder-27k\cite{liu2023rtlcoder}. \iffalse ITERTL-DeepSeekV2-7k*, which was trained using code generated by GPT-4o, is placed in the third group instead of the second to avoid exaggerating the advantages provided by the data.\fi Within each group with reported data volume, the best results are marked in \textbf{bold}. Some results are missing because they were not reported in the corresponding papers, and both models and training data are closed-source, making it impossible to reproduce.\iffalse  If the best result comes from a closed-source model (like GPT-4), then the best result from the open-source model is also highlighted in bold. Additionally, the second best result is underlined.\fi}
  \label{tab:comp}
  \begin{tabular}{ccccccccc}
    \toprule
    \multirow{2}{*}{Group}&\multirow{2}{*}{Model}&\multirow{2}{*}{Num}&\multicolumn{3}{c}{VerilogEval-Machine(\%)}&\multicolumn{3}{c}{VerilogEval-Human(\%)}\\ \cline{4-9}
                      & &  &pass@1&Pass@5&pass@10&pass@1&Pass@5&pass@10\\
    \midrule
    % GPT-3.5 (Closed-Source) &- &46.7& 69.1& \textbf{74.1}& 26.7& 45.8& 51.7\\
    % GPT-4 (Closed-Source) &- &\textbf{60.0}& \textbf{70.6}& 73.5& \textbf{43.5}& \textbf{55.8}& \textbf{58.9}\\

    % Codegen\cite{nijkamp2022codegen} &- &5.0 &17.6 &25.8 &1.6 &6.1 &9.4\\
    % DeepSeek-Coder\cite{guo2024deepseek} &- &52.2 &55.4 &56.8 &30.2 &33.9 &34.9 \\
    % \hdashline
    % VeriGen\cite{thakur2023benchmarking} &-  &33.8& 59.2& 67.9& 24.5& 45.3& 53.2\\
    % % BetterV-CodeQwen\cite{pei2024betterv} &-  &\textbf{68.1} &\textbf{79.4} &\textbf{84.5} &46.1 &53.7 &58.2\\
    % BetterV-DeepSeek\cite{pei2024betterv} &- &\textbf{67.8} &\textbf{79.1} &\textbf{84.0} &45.9 &53.3 &57.6 \\
    % ChipNeMo\cite{liu2023chipnemo} &- &43.4 &- &- &22.4 &- &-\\
    % % CodeGen-6B MEV-LLM\cite{nadimi2024multi} &- &57.3 &61.5 &66.4 &42.9 &48 &54.4\\
    % CodeGen-16B MEV-LLM\cite{nadimi2024multi} &- &60.8 &63.6 &69.2 &\textbf{47.4} &\textbf{55.7} &\textbf{60.2}\\

    % \hdashline

    % no bold for model without data
    \multirow{7}{*}{\makecell{Foundation Models or \\ Data Volume Not Reported}}& GPT-3.5 (Closed-Source) &- &46.7& 69.1& 74.1& 26.7& 45.8& 51.7\\
    &GPT-4 (Closed-Source) &- &60.0& 70.6& 73.5& 43.5& 55.8& 58.9\\

    % Codegen\cite{nijkamp2022codegen} &- &5.0 &17.6 &25.8 &1.6 &6.1 &9.4\\
    % &DeepSeek-Coder\cite{guo2024deepseek} &- &52.2 &55.4 &56.8 &30.2 &33.9 &34.9 \\
    % DeepSeek-Coder-V2\cite{zhu2024deepseek} &- &72.7 &83.8 &87.4 &48.7 &60.4 &63.5 \\
    % \hdashline
    &VeriGen\cite{thakur2023benchmarking} &-  &33.8& 59.2& 67.9& 24.5& 45.9& 53.2\\
    % BetterV-CodeQwen\cite{pei2024betterv} &-  &\textbf{68.1} &\textbf{79.4} &\textbf{84.5} &46.1 &53.7 &58.2\\
    &BetterV-DeepSeek\cite{pei2024betterv} &- &67.8 &79.1 &84.0 &45.9 &53.3 &57.6 \\
    &ChipNeMo\cite{liu2023chipnemo} &- &43.4 &- &- &22.4 &- &-\\
    % CodeGen-6B MEV-LLM\cite{nadimi2024multi} &- &57.3 &61.5 &66.4 &42.9 &48 &54.4\\
    % &CodeGen-16B MEV-LLM\cite{nadimi2024multi} &- &60.8 &63.6 &69.2 &47.4 &55.7 &60.2\\

    \midrule

    \multirow{4}{*}{Data Volumn $\leq$ 10k}&VerilogEval-CodeGen\cite{liu2023verilogeval} &8.5k  &46.2 &67.3 &73.7 &28.8 &45.9 &52.3\\
    &RTLCoder-DeepSeek-10k\cite{liu2023rtlcoder} &10k &55.3 &70.4 &76.2 &36.7 &47.0 &50.4\\
    &\textbf{ITERTL-DeepSeek-8k} &8k  &60.8 &\textbf{74.7} &\textbf{81.1} &41.7 &\textbf{49.7} &\textbf{52.6}\\
    &\textbf{ITERTL-DeepSeek-10k} &10k  &\textbf{61.5} &73.0 &76.9 &\textbf{42.3} &49.4 &\textbf{52.6}\\

    % \hdashline
    \midrule

    \multirow{2}{*}{10k$ \textless $Data Volumn $ \leq $27k}&RTLCoder-DeepSeek-27k\cite{liu2023rtlcoder}  &27k &\textbf{61.2} &\textbf{76.5} &\textbf{81.8} &41.6 &\textbf{50.1} &53.4\\
    % CodeV-DeepSeek\cite{zhao2024codev} &165k  &77.9 &88.6 &90.7 &52.7 &62.5 &67.3\\
    % CodeV-CodeQwen\cite{zhao2024codev} &165k  &77.6 &88.2 &90.7 &53.2 &65.1 &68.5\\
    
    % &\textbf{ITERTL-DeepSeek-22k} &22k  &60.8 &73.6 &77.6 &\textbf{45.5} &\textbf{50.1} &\textbf{55.8}\\
    &\textbf{ITERTL-DeepSeek-22k} &22k  &60.8 &73.6 &77.6 &\textbf{45.5} &\textbf{50.1} &\textbf{55.8}\\

    % \hdashline
    \midrule

    \multirow{2}{*}{Higher Quality Data}&CodeV-DeepSeek-27k*\cite{zhao2024codev} &27k*  &\textbf{72.3} &\textbf{82.2} &\textbf{85.6} &39.1 &53.4 &58.3\\
    &\textbf{ITERTL-DeepSeekV2-7k*} &7k*  &61.5 &71.4 &75.5 &\textbf{53.8} &\textbf{60.8} &\textbf{64.1}\\

  \bottomrule
\end{tabular}
%\vspace{-12pt}
\end{table*}

\textbf{Metric:} As a widely used metric, pass@$k$ metric \cite{kulal2019spoc} is employed to measure the performance, by regarding a problem as solved if at least one of the $k$ generated code samples passes the unit tests. \iffalse Specifically, for each problem, the model generates $n\ge k$ candidates (n = 10 in our experiments.), where $c\le n$ samples pass the unit tests. The pass@$k$ metric is estimated unbiasedly using the following expression \cite{chen2021evaluating}:

\begin{equation}
pass@ k=\underset{Problems}{\mathbb{E}}\left[1-\frac{\binom{n-c}{k}}{\binom{n}{k}}\right]
\end{equation}\fi

% For a comprehensive evaluation, we measure the pass@$k=\{1,5,10\}$ metrics setting $n=10$ in main results. To avoid the impact of randomness and improve efficiency, we focus more on the pass@1 value under greedy decoding in other experiments. 

\textbf{Decoding Strategy:} 
During the training stage, in order to enhance the diversity and promote the exploration, we use Top-p decoding strategy with $top_{p}=0.95$ and $temperature = 0.5$.
In the testing stage, to assess performance comprehensively, the model is prompted to generate responses with $temperature = \{0(greedy\, decoding),0.2,0.5,0.8\}$ and $top_{p}=0.95$. For the most representative metric pass@1, we directly use the result from greedy decoding to avoid random fluctuations. For other test metrics, the best result from different temperatures is selected.

\textbf{Training Details:} 
DeepSeek-Coder-Instruct-6.7B \cite{guo2024deepseek} and DeepSeek-Coder-V2-Lite-Instruct-16B \cite{zhu2024deepseek} are chosen as base models. We employ nearly 27k open-source instruction-code pairs from 
RTLCoder \cite{liu2023rtlcoder} as reference data. \iffalse This dataset contains nearly 27k samples generated with GPT-3.5. Datasets of different scales are construct by randomly sampling and filtering from the original dataset.  For the final version of our SOTA models, we trained a 6.7B model with about 22k samples, and a 16B model with about 7k samples due to computational limitations. \fi We also regenerate the corresponding code using GPT-4o based on original instructions to construct higher quality reference data, which is used to fine-tune ITERTL-DeepSeekV2-7k*. \iffalse The learning rate is set to $10^{-5} $. AdamW optimizer is employed with $\beta _{1} =0.9$ and $\beta _{2} =0.999$.  The bp16 Mixed-precision Training is adopted to avoid overflow. During each iteration, the model is trained for 3 epochs.\fi The total number of iterations is set to 7. In actual applications, we decide whether to continue the iteration based on whether the loss continues to decrease significantly, or simply select the number of iterations to be 3, which usually is sufficient based on our experience. \iffalse We conduct all experiments on NVIDIA A800 GPUs.\fi The number of candidate responses $K$ is set to 4. The hyper-parameters $\alpha$ and $\beta $ in Equation~\ref{eqn:ranking} are set to 0.3 and 0.2, respectively. 

%\vspace{-5pt}
\subsection{ Main Results}

% As Table~\ref{tab:comp} shows, our model reach the state-of-the-art level among open-source models. On pass@1 metric, our model trained with only 10k reference samples surpasses RTLCoder with 27k reference samples by 1\% and 1.3\% on VerilogEval-machine and VerilogEval-human respectively. Leveraging an equal amount of reference samples, our model significantly outperforms RTLCoder-DeepSeek-10k on all metrics, with a
% relative improvement of 12.5\% and 16.9\% in pass@1 especially. Considering the size of the parameters, with only 6.7 billion parameters, our model significantly outperforms VerilogEval \cite{liu2023verilogeval} with 16 billion parameters and a similar volume of reference samples (8502), further proving the efficiency and superiority of our method. Additionally, we incorporated a general-purpose code model, Codegen \cite{nijkamp2022codegen}, into our comparison. Its subpar performance on the VerilogEval benchmark underscores the importance of tailoring LLMs for RTL code.

% Relative to closed-source models, our model comprehensively surpasses GPT-3.5 on the benchmark. Against GPT-4, our model performs better on VerilogEval-machine but lags on VerilogEval-human. Considering the vast differences in training costs and the lightweight advantage of our model, these performance deficits are acceptable.

As Table~\ref{tab:comp} shows, our model reaches the SOTA level in benchmarks. With just 7k annotated reference data, our model ITERTL-DeepSeekV2-7k* surpasses GPT-4 and outperforms all other models in the table on every metric of the Verilog-Human benchmark, achieving an outstanding 53.8\% pass@1. On the VerilogEval-Machine benchmark, our four models (ITERTL-DeepSeek-8k, ITERTL-DeepSeek-10k, ITERTL-DeepSeek-22k, ITERTL-DeepSeekV2-7k*) outperform GPT-4 across all metrics, yet remain behind CodeV-DeepSeek-27k*\cite{zhao2024codev}. 

However, it is important to note that CodeV employs a larger quantity of high-quality data compared to us. These data haven't yet been publicly released, making it impossible for us to conduct a fair comparison. What's more, as noted in the VerilogEval\cite{liu2023verilogeval},  instructions generated by LLMs for the Verilog-Machine are often verbose and unfocused on core functionalities, with the potential for ambiguities and errors due to lack of thorough review.   In contrast, the Verilog-Human benchmark  provides more representative results, highlighting the superiority of our approach.

By controlling for a similar number and quality of reference samples, our models demonstrate a significant advantage comparing with baselines. ITERTL-DeepSeek-10k outperforms RTLCoder-DeepSeek-10k across all metrics, achieving a relative improvement of 15.3\% in the pass@1 metric on Verilog-Human benchmark. ITERTL-DeepSeek-22k significantly surpasses RTLCoder-DeepSeek-27k on Verilog-Human benchmark ,with a 9.3\% relative enhancement in the pass@1 metric. Overall, under the same conditions of reference data volume and quality, our approach significantly surpasses the baseline method. \iffalse Furthermore, by incorporating higher-quality data, our model achieves state-of-the-art performance, which could be further enhanced with the addition of more new high-quality data in the future.\fi

% On pass@1 metric, our model trained with only 10k reference samples surpasses RTLCoder with 27k reference samples by 1\% and 1.3\% on VerilogEval-machine and VerilogEval-human respectively. Leveraging an equal amount of reference samples, our model significantly outperforms RTLCoder-DeepSeek-10k on all metrics, with a
% relative improvement of 12.5\% and 16.9\% in pass@1 especially. Considering the size of the parameters, with only 6.7 billion parameters, our model significantly outperforms VerilogEval \cite{liu2023verilogeval} with 16 billion parameters and a similar volume of reference samples (8502), further proving the efficiency and superiority of our method. Additionally, we incorporated a general-purpose code model, Codegen \cite{nijkamp2022codegen}, into our comparison. Its subpar performance on the VerilogEval benchmark underscores the importance of tailoring LLMs for RTL code.

% Relative to closed-source models, our model comprehensively surpasses GPT-3.5 on the benchmark. Against GPT-4, our model performs better on VerilogEval-machine but lags on VerilogEval-human. Considering the vast differences in training costs and the lightweight advantage of our model, these performance deficits are acceptable.

%\vspace{-5pt}
\subsection{Effect of Iterations}

% \begin{figure}[!h]
% %\vspace{-8pt}
%   \centering
%   \includegraphics[width=.8\linewidth]{fig_comp_v4_c3_crop.pdf}
%   \caption{Number of Iterations and pass@1 on VerilogEval-human. As the iteration count increases, the pass@1 rate approximately rises initially and then decreases due to the overfitting.}
%   \label{fig:comp}
% %\vspace{-15pt}  
% \end{figure}
% %\vspace{-8pt}

% To delve deeper into the impact of iteration, we plot how pass@1 on VerilogEval-human varies with the number of iterations in Figure~\ref{fig:comp}. It can be clearly seen that, with 2 iterations and only 10k reference data, our model outperforms the baseline model with 27k reference data. Furthermore, our model outperforms GPT4 in pass@1 metric with 22k reference data. In the early stages of iterations, the pass@1 rate exhibits a clear upward trend, indicating the efficacy of iterative training. After reaching the peak, the pass@1 rate begins to decline. This trend suggests an overfitting, which can be avoid by early stopping. \iffalse In fact, similar discoveries were observed on LLMs trained using reward models on natural languages \cite{gao2023scaling}. Our work reveals that LLMs trained based on RTL code quality evaluation have similar properties, providing inspiration for subsequent improvement work.\fi 

\begin{figure}[!h]
  \centering
  \includegraphics[width=.8\linewidth]{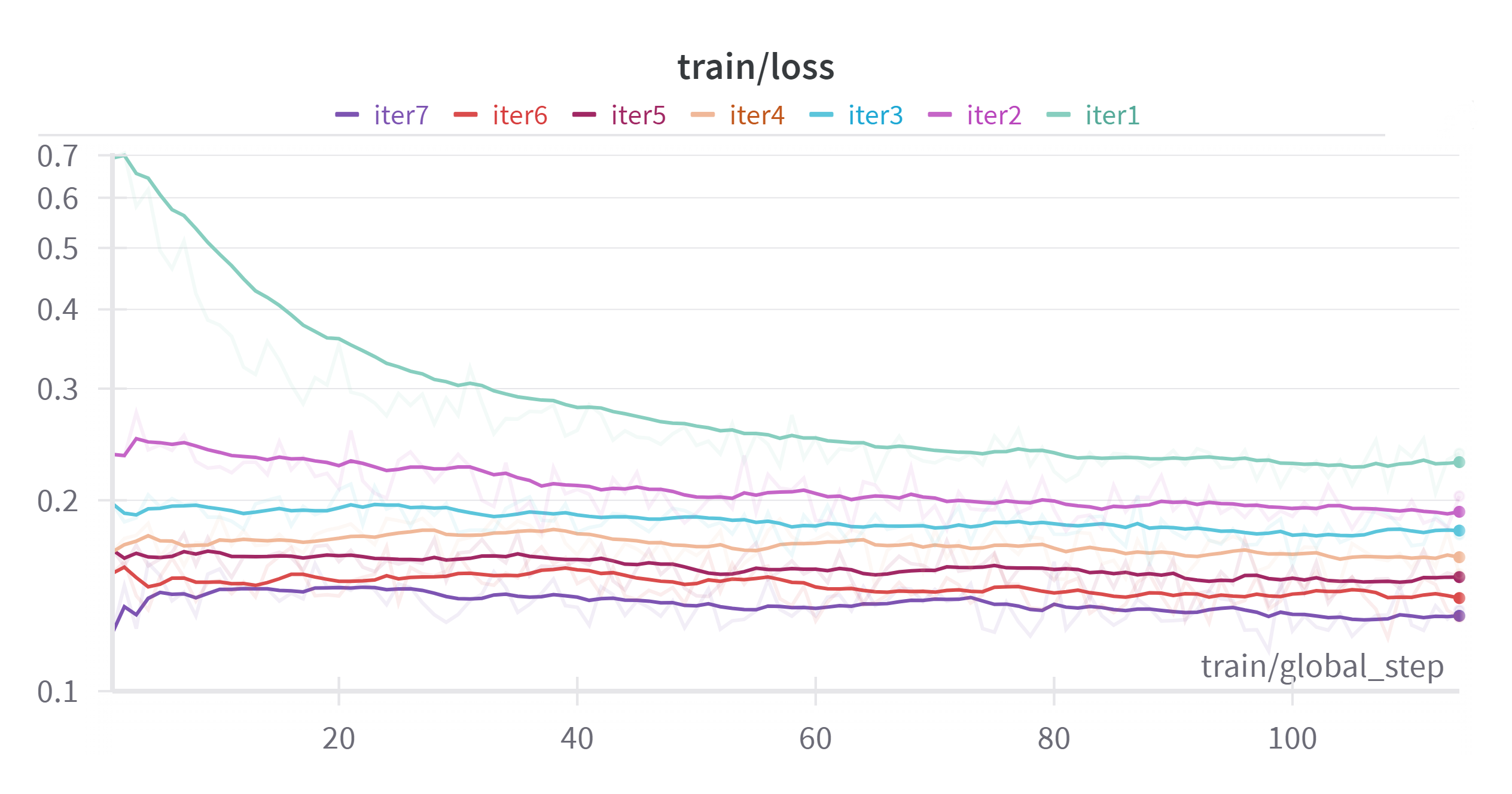}
  \caption{The loss function curves across each iteration. For better visualization, we represent the original loss function curves with light-colored lines and the results of exponential smoothing with dark-colored lines. And the vertical axis is on a log scale. As the iteration count increases, the loss function decreases.}
  \label{fig:loss}
%\vspace{-13pt}  
\end{figure}

Figure~\ref{fig:loss} depicts different loss functions curves across each iteration. Even though the loss function converges within each round, the loss value can still decrease by leveraging newly sampled data from updated model, validating the effectiveness of the proposed iterative training approach. Another observed phenomenon is that as the number of iterations increases, the marginal decrease in the loss function gradually diminishes\iffalse (the vertical axis is on a log scale)\fi, which indicates that the loss function is nearing its optimization ceiling.

% %\vspace{-8pt}

\begin{figure}[!htb]
\begin{minipage}[t]{.48\linewidth}
    \centering
    \subfloat[][Incorrect code]{\label{incorrect, non-self-contained code}\includegraphics[height=2.5cm]{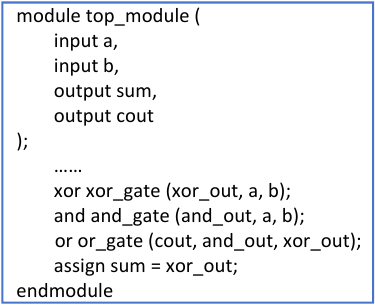}}

\end{minipage} %\par
% \medskip
% \\
\begin{minipage}[t]{.48\linewidth}
    \centering

    \subfloat[][Correct code]{\label{Finetune with reference data and sampled data}\includegraphics[height=2.5cm]{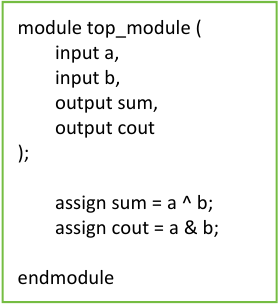}}
\end{minipage} %\par

\caption{Examples of output code for a half adder. (a) shows the error implementation generated by model trained without iteration. (b) shows the correct implementation generated by model trained with iteration.}
\label{fig:half_adder}
\end{figure}

To better demonstrate the effects of iteration, we present the implementation of a half adder using LLMs trained with and without iteration. As Figure~\ref{fig:half_adder} (a) shows, the model trained without iteration generates wrong code which incorrectly defines the carry-out (cout) using an OR gate. As a comparison, the model, after iterative training, produces the correct implementation in Figure~\ref{fig:half_adder} (b), using an AND gate to generate the cout value.

% Figure~\ref{fig:loss} depicts different loss functions curves across each iteration. Even though the loss function converges within each round, the loss value can still decrease by leveraging newly sampled data from updated model, validating the effectiveness of the proposed iterative training approach. Another observed phenomenon is that as the number of iterations increases, the marginal decrease in the loss function gradually diminishes\iffalse (the vertical axis is on a log scale)\fi, which indicates that the surrogate reward function is nearing its optimization ceiling. 

% %\vspace{-8pt}
% \begin{figure}[!htbp]
%   \centering
%   \includegraphics[width=.8\linewidth]{loss_log.png}
%   % \includegraphics[width=.8\linewidth]{loss_iter0816.png}
%   \caption{The loss function curves across each iteration. For better visualization, we represent the original loss function curves with light-colored lines and the results of exponential smoothing with dark-colored lines. And the vertical axis is on a log scale. As the iteration count increases, the loss function decreases.}
%   \label{fig:loss}
% %\vspace{-13pt}  
% \end{figure}

% %\vspace{-5pt}
% %\vspace{-5pt}
\subsection{Ablation Study of Data Filter}

% \subsubsection{Impact of Data Filter}

% \begin{table}[!htbp]
% 	\centering        % 表格居中
	
% 	\caption{pass@1 on VerilogEval-human of approaches with and without data filter.} % 大括号内定义图片标题
%         \label{tab:ab_df} % 大括号内定义图片标签，用于正文引用
% 	%\vspace{2pt}      % 大括号内设置表格与正文之间的间距
% 	\begin{tabular}{c|c|c} % tabular定义了表格本身 {l|cc|cc}定义了表格共有6列，以及每一列的对齐方式（l左对齐，c居中，r右对齐），且第一列和第二、三列和第四、五列之间用竖线隔开。如果是三列中且没有竖线隔开就是{cccc}。
% 		\hline % 定义表格的横线
% 		~ &  Non-iterative & Iterative \\ \hline 
% 		Without Data Filter  & 42.9 & 51.9  \\ \hline % 定义每一行单元格内容，其中，第一个单元格前不加“&”，其他每个单元格前加“&”，“\\”表示换行
		
% 		With Data Filter  &\textbf{51.9} &\textbf{53.8} \\ \hline
% 	\end{tabular}
% \end{table}
% %\vspace{-8pt}

To validate the effectiveness of the proposed data filter, we conduct the ablation experiment on the DeepSeek-Coder-V2-Lite-Instruct-16B \cite{zhu2024deepseek} model using 10k samples, with the code  regenerated by GPT-4o. As Table~\ref{tab:ab_df} shows, with the combination of the data filter and the iterative training scheme, our model achieves a pass@1 score of 53.8\%, significantly outperforming the 51.9\% score obtained without the data filter. \iffalse Additionally, the model reached its performance peak earlier, in just 2 iterations compared to 3 iterations without the data filter. \fi \iffalse This observation suggests that the data filter effectively enhances the iterative training process, resulting in improved performance.\fi When considering the model's performance without iterative training, we find that the data filtering mechanism is still effective,  boosting the pass@1 score from 42.9\% to 51.9\%. \iffalse Surprisingly, if there is no data filtering mechanism, the model's performance actually declined after the first iteration. This could be attributed to the unfiltered reference code being excessively verbose, leading to overly updates  to the model's parameters. \fi The above analysis demonstrates that the proposed data filter, as a plug-and-play strategy, can either be used independently to enhance performance or be combined with the iterative training scheme to further unlock models' capability.

%\vspace{-5pt}
\begin{table}[!htbp]

	\centering        % 表格居中
	
	\caption{pass@1 on VerilogEval-human of approaches with and without data filter.} % 大括号内定义图片标题
        \label{tab:ab_df} % 大括号内定义图片标签，用于正文引用
	%\vspace{2pt}      % 大括号内设置表格与正文之间的间距
	\begin{tabular}{c|c|c} % tabular定义了表格本身 {l|cc|cc}定义了表格共有6列，以及每一列的对齐方式（l左对齐，c居中，r右对齐），且第一列和第二、三列和第四、五列之间用竖线隔开。如果是三列中且没有竖线隔开就是{cccc}。
		\hline % 定义表格的横线
		~ &  Non-iterative & Iterative \\ \hline 
		Without Data Filter  & 42.9 & 51.9  \\ \hline % 定义每一行单元格内容，其中，第一个单元格前不加“&”，其他每个单元格前加“&”，“\\”表示换行
		
		With Data Filter  &\textbf{51.9} &\textbf{53.8} \\ \hline
	\end{tabular}
%\vspace{-5pt}
\end{table}

\section{Conclusion}

In this paper, we propose an iterative training paradigm to fine-tune LLMs for RTL code generation. During each iteration, the model trained in the previous iteration is employed to update samples, which are then utilized for training in the current round. \iffalse From a perspective of maximizing the reward function, we theoretically analyze the superiority of our approach.\fi Furthermore, we propose a data filter to improve the accuracy of RTL code generation. Experiments demonstrate the data efficiency of our method and the superior functional correctness of the generated RTL code. \iffalse With just about 7k high-quality reference samples, our model outperforms GPT-4 and state-of-the-art open-source LLMs with 53.8\% pass@1 rate on VerilogEval-human benchmark. While maintaining similar data quality and quantity, our models demonstrate relative performance improvements of 9.3\% and 15.3\% over the non-iterative baseline with
data sizes of 27k and 10k.\fi  Our framework has the potential to further assist hardware design by incorporating more tools in the future.

\section*{Acknowledgment}
This work was supported by Beijing Natural Science Foundation (Grant No.L234078) and Beijing Nova Program (Grant No. 20220484054 and No. 20230484420).

% \appendix
\bibliographystyle{IEEEtran}
\bibliography{IEEEabrv,aaai25}

\end{document}